\documentclass[10pt,twocolumn,letterpaper]{article}

\usepackage{iccv}
\usepackage{times}
\usepackage{epsfig}
\usepackage{graphicx}
\usepackage{amsmath}
\usepackage{amssymb}	
\usepackage{xcolor}
\usepackage{caption}
\usepackage{balance}
\usepackage{booktabs}

\usepackage[pagebackref=true,breaklinks=true,letterpaper=true,colorlinks,bookmarks=false]{hyperref}

\iccvfinalcopy 

\def\iccvPaperID{****} 

\ificcvfinal\fi

\begin{document}

\title{Probabilistic Modeling for Human Mesh Recovery}
\author{Nikos Kolotouros$^1$, Georgios Pavlakos	$^2$, Dinesh Jayaraman $^1$ Kostas Daniilidis$^1$ \\[0ex]
	$^1$ University of Pennsylvania \hspace{0.1em} $^2$ UC Berkeley
}

\ificcvfinal\thispagestyle{empty}\fi

\maketitle

\begin{abstract}
This paper focuses on the problem of 3D human reconstruction from 2D evidence. Although this is an inherently ambiguous problem, the majority of recent works avoid the uncertainty modeling and typically regress a single estimate for a given input. In contrast to that, in this work, we propose to embrace the reconstruction ambiguity and we recast the problem as learning a mapping from the input to a \textbf{distribution} of plausible 3D poses. Our approach is based on the normalizing flows model and offers a series of advantages. For conventional applications, where a single 3D estimate is required, our formulation allows for efficient mode computation. Using the mode leads to performance that is comparable with the state of the art among deterministic unimodal regression models. Simultaneously, since we have access to the likelihood of each sample, we demonstrate that our model is useful in a series of downstream tasks, where we leverage the probabilistic nature of the prediction as a tool for more accurate estimation. These tasks include reconstruction from multiple uncalibrated views, as well as human model fitting, where our model acts as a powerful image-based prior for mesh recovery. Our results validate the importance of probabilistic modeling, and indicate state-of-the-art performance across a variety of settings. Code and models are available at: \url{https://www.seas.upenn.edu/~nkolot/projects/prohmr}.
\end{abstract}

\section{Introduction}
\begin{figure}[!t]
\includegraphics[width=\columnwidth,trim={0 2cm 11cm 0},clip]{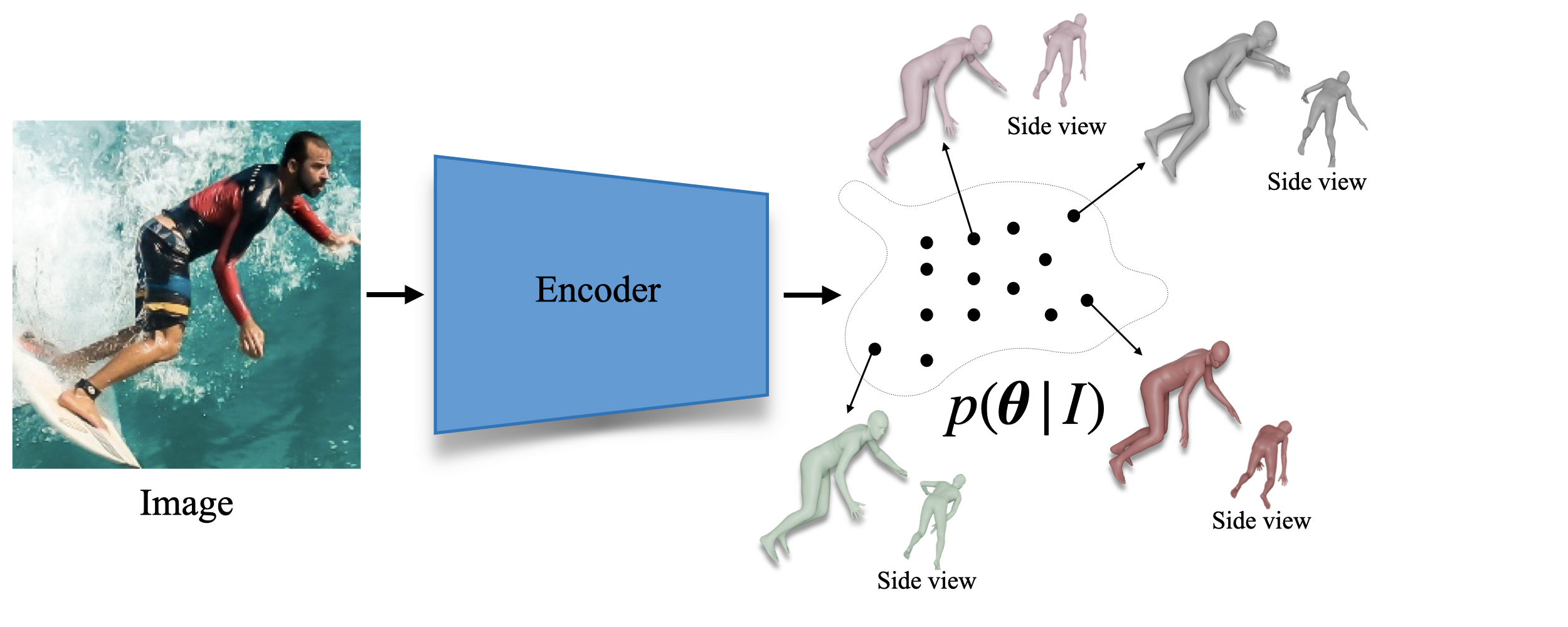} 
\caption{\textbf{Probabilistic modeling for 3D human mesh recovery}. 
We propose to recast the problem of 3D human reconstruction as learning a mapping from the input to a distribution of 3D poses. The output distribution has high probability mass on a diverse set of poses that are consistent with the 2D evidence.
}
\vspace{-0.1in}
\label{fig:teaser}
\vspace{-0.15in}
\end{figure}

\begin{figure*}[!htb]
\vspace{-0.2in}	
	\includegraphics[width=\textwidth,clip]{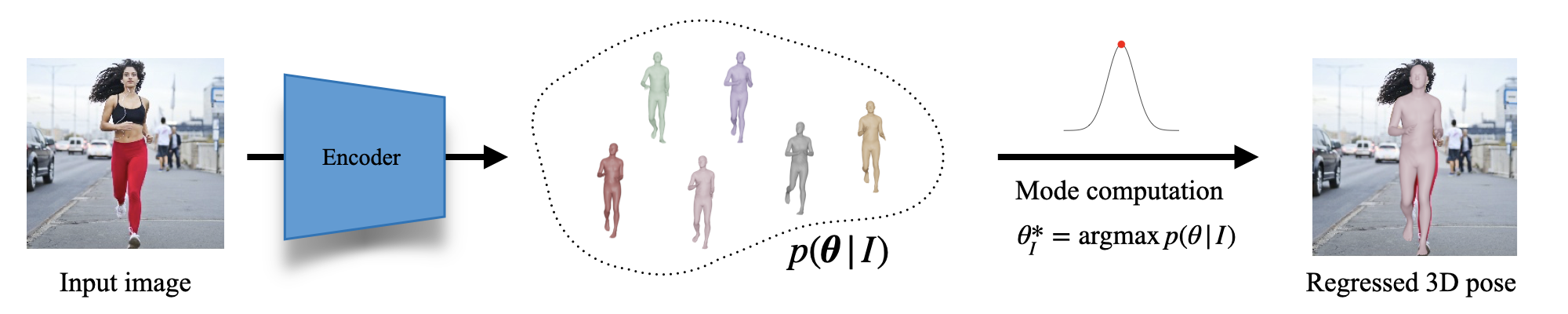}\\
	\vspace{-0.05in}
	\includegraphics[width=\textwidth,clip]{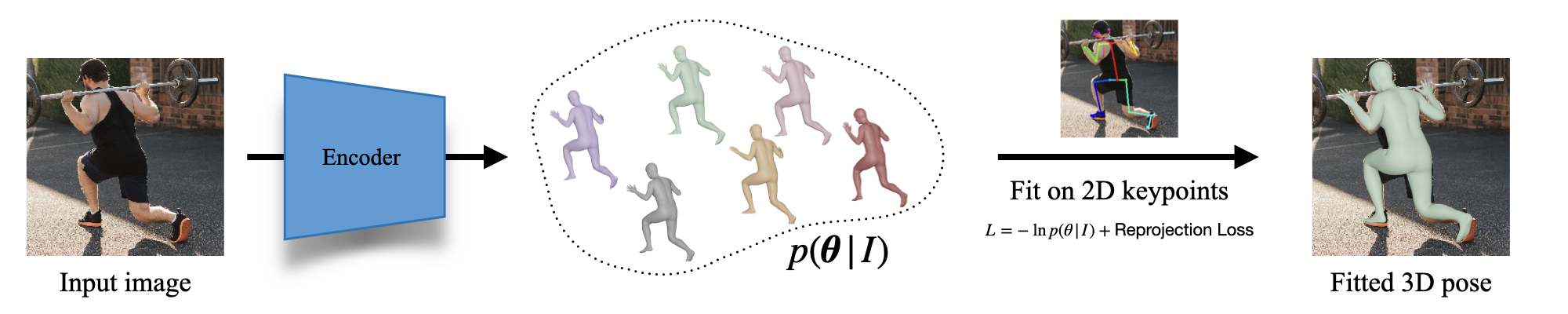}\\
	\vspace{-0.1in}
	\includegraphics[width=\textwidth,clip]{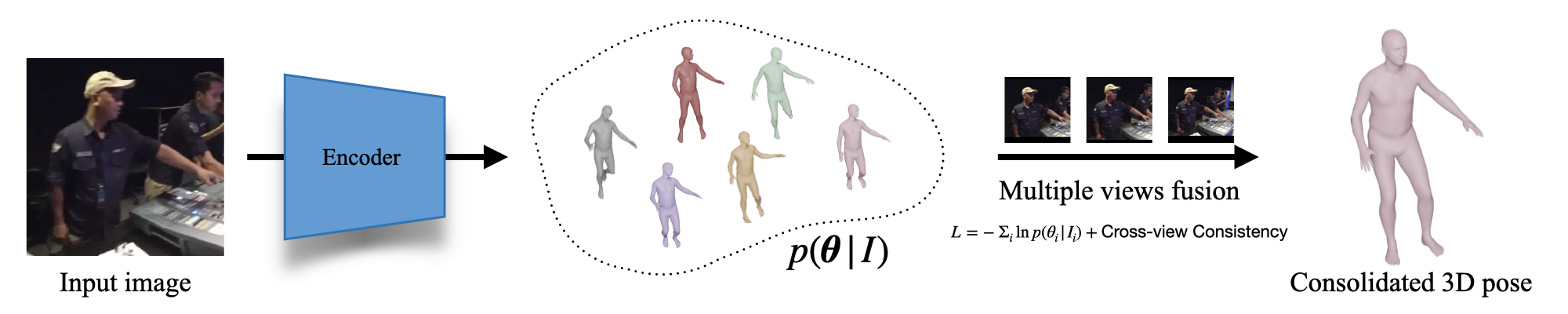}
	\vspace{-0.15in}
	\caption{\textbf{The value of probabilistic modeling for 3D human mesh estimation}. We demonstrate that probabilistic modeling in the case of 3D human mesh estimation can be particularly useful because of its elegant and flexible form, which enables a series of downstream applications. First row: In the typical case of 3D mesh regression, we can naturally use the mode of the distribution and perform on par with approaches regressing a single 3D mesh. Second row: When keypoints (or other types of 2D evidence) are available we can treat our model as an image-based prior and fit a human body model to the keypoints by combining it with a 2D reprojection term. Third row: When multiple views are available, we can naturally consolidate all single-frame predictions by adding a cross-view consistency term. We underline that all these applications refer to test-time behavior and they use the same trained probabilistic model (no per-task training required).}
	\vspace{-0.15in}
	\label{fig:downstream}
\end{figure*}
Reconstructing 3D human pose from any form of 2D observations (image, 2D keypoints, silhouettes) is a fundamentally ambiguous problem. Of course, this is a very old insight, identified even from the very first approaches~\cite{lee1985determination} dealing with the problem of single-view human pose reconstruction. However, the current norm for the state-of-the-art approaches is to return a single 3D estimate which is typically computed in a deterministic manner. In this work, we argue that there is great value at capturing a distribution of 3D poses conditioned on the preferred input.

Our reliance on systems that return a single deterministic 3D pose output often happens out of convenience; it makes comparison on conventional benchmarks straightforward and fair, while a single output is enough for many downstream applications. Recent literature for 3D human pose reconstruction is currently dominated by such approaches and they are very popular for image~\cite{kolotouros2019spin} or keypoint~\cite{song2020human} input, for skeleton-based~\cite{martinez2017simple} or mesh-based~\cite{kolotouros2019convolutional} reconstruction, as well as regression~\cite{kanazawa2018end} or optimization-based~\cite{bogo2016keep} approaches. On the other end of the spectrum, there have always been approaches that advocate in favor of generating multiple predictions per input. Recent efforts have demonstrated interesting potential~\cite{biggs2020multibodies, li2019cvpr}, but often rely on ensemble-type prediction, modifying current systems into combining $N$ output heads instead of one. This can lead to cumbersome architectural choices, inability to scale and/or limited expressivity for the output distribution.

Our approach aims to bridge this gap and demonstrate the value of predicting a distribution of 3D poses conditioned on the provided 2D input. To achieve this, we propose an elegant and efficient approach with many desirable properties missing from recent work, and we demonstrate its effectiveness. Instead of regressing a single estimate for the provided input, we use Normalizing Flows to regress a distribution of plausible poses. This allows us to train a network which returns a conditional distribution of 3D poses as a function of the input (\eg, image or 2D keypoints), as depicted in Figure~\ref{fig:teaser}. Our probabilistic model allows for fast sampling of diverse outputs, we can efficiently compute the likelihood of each sample, and there is a fast and closed form solution to compute the mode of the distribution. The importance of the above is manifested in a variety of ways, which are summarized in Figure~\ref{fig:downstream}. First, we can easily compute the mode of the distribution, which returns the most likely 3D pose for the particular input. This is convenient, when a single estimate is required for some applications. Interestingly, this regressed value is on par with the state-of-the-art deterministic methods, so our model can be valuable even in the more conventional settings. More importantly though,  by treating our trained probabilistic model as a conditional distribution, we can use it in many downstream applications to combine information from different sources. For example, when 2D keypoints are available, optimization approaches~\cite{bogo2016keep, pavlakos2019expressive}, are used to fit parametric human body models to these 2D locations. In this case, our model can act as a powerful image-based prior that can guide the optimization towards accurate solutions that satisfy both 2D keypoint reprojection and image evidence. Similarly, when multiple views are available, we can consolidate information from all conditional distributions, by optimizing for cross-view consistency and recover a 3D result that is consistent with the available observations. Last but not least, we highlight that all these applications are available at test-time with the same trained probabilistic model, without any need for task-specific retraining.

We conduct extensive experiments to demonstrate the importance of our learned probabilistic model. We focus primarily on image-based mesh recovery~\cite{kanazawa2018end}, proposing the \textbf{ProHMR} model, but we also investigate 2D keypoint input~\cite{martinez2017simple}. We achieve particularly strong performance across different tasks and evaluation settings. Our contributions can be summarized as follows:
\begin{itemize}
    \item We propose a probabilistic model for human mesh recovery and demonstrate its value in various tasks.
    \item In the conventional evaluations with single estimate methods, our model is on par with the state of the art.
    \item We demonstrate that in the presence of additional information sources, \eg, multiple views or 2D keypoints, our model offers an elegant and effective way to consolidate said sources.
    \item In the setting of human body model fitting, our model acts as a powerful image-based prior, achieving significant boost over previous baselines.
\end{itemize}

\section{Related work}
Although our formulation is quite general and can handle different inputs/outputs, here we focus mainly on human mesh recovery from a single image~\cite{kanazawa2018end}, while we briefly touch upon other settings, specifically 3D pose estimation from 2D keypoints~\cite{martinez2017simple}. Since the related work is vast, here we discuss the more relevant approaches. We direct the interested reader to a recent and extensive survey~\cite{zheng2020deep}.

\subsection{Human mesh recovery from a single image}

\noindent
\textbf{Regression:}
Recent approaches for mesh recovery are following the regression paradigm, where the parameters of a parametric model~\cite{loper2015smpl, pavlakos2019expressive, xu2020ghum, osman2020star} are regressed from a deep network, given a single image as input. The canonical example here is HMR~\cite{kanazawa2018end}, with many of the design decisions being adopted also by follow-up works, \eg,~\cite{arnab2019exploiting,guler2019holopose, kolotouros2019convolutional,pavlakos2019texturepose, choutas2020monocular, georgakis2020hierarchical, jiang2020multiperson}. Here, our regression network also follows the principles of HMR, however, instead of regressing a single 3D pose estimate, it regresses a whole distribution of plausible 3D poses given the input image.

\noindent
\textbf{Optimization:}
These methods estimate iteratively the parameters of the body model, such that it is consistent with a set of 2D cues. The canonical example of SMPLify~\cite{bogo2016keep} optimizes SMPL parameters given 2D keypoints. Follow-up works investigate other inputs, \eg, silhouettes~\cite{lassner2017unite}, POFs~\cite{xiang2019monocular},  dense correspondences~\cite{guler2019holopose} or contact~\cite{muller2021self, taheri2020grab}. However,  most recent approaches~\cite{arnab2019exploiting, kolotouros2019spin,pavlakos2019expressive} rely almost exclusively on 2D keypoints; losing the majority of pictorial cues,  but gaining robustness. In this work, we demonstrate how our probabilistic model can leverage image-based information to guide the keypoint-based optimization.

\noindent
\textbf{Optimization-Regression hybrids:}
The idea of building a hybrid between the two paradigms has been explored extensively in recent work. HMR~\cite{kanazawa2018end} and HUND~\cite{zanfir2020neural} use a network to mimic the optimization steps and regress the updates to the model parameters. Song~\etal~\cite{song2020human} use the reprojection error of the model joints to guide their learning-based gradient descent approach. SPIN~\cite{kolotouros2019spin} initializes the optimization with a regression network and supervises the network with the output of the optimization. EFT~\cite{joo2020eft} builds on that by updating the network weights during the fitting procedure. Our probabilistic model also investigates this type of collaboration by regressing a distribution of poses which can then be used as a prior term for the fitting.

\subsection{Multiple hypotheses for 3D human pose}

Multiple hypotheses methods have been used in the context of 3D human pose estimation to deal with the inherent ambiguities of the reconstruction such as occlusions, truncations or depth ambiguities. Jahangiri and Yuille~\cite{jahangiri2017generating} use a compositional model and anatomical constraints to generate multiple hypotheses consistent with 2D keypoint evidence. Li and Lee~\cite{li2019cvpr} use a Mixture Density Network instead and generate a fixed number of proposals based on the centroids of the Gaussian kernels, while Sharma~\etal~\cite{sharma19monocular} tackle the same problem using a Conditional VAE. Recently, Biggs~\etal~\cite{biggs2020multibodies} extend HMR~\cite{kanazawa2018learning} with $N$ prediction heads.  This leads to a discrete set of hypotheses, instead of a full probability of poses as we do. In a concurrent work, Sengupta~\etal~\cite{sengupta2021probabilistic} use a Gaussian posterior to model the uncertainty in the parameter prediction. Differently from these methods, our approach is not limited to learning a generative model of plausible 3D poses, but rather shows how one can use such a model for useful downstream applications.

\subsection{Normalizing Flows}
Normalizing Flows are used to represent complex distributions as a series of invertible transformations of a simple base distribution.
They were originally developed for modeling posterior distributions for variational inference~\cite{rezende15flows, kingma2016improved}.
Popular examples include MADE~\cite{germain15maf}, NICE~\cite{dinh2015nice}, MAF~\cite{papamakarios17maf}, RealNVP~\cite{dinh2017realnvp} and Glow~\cite{kingma2018glow}.

Normalizing Flows have been used in the context of 3D human pose estimation to learn a prior on the distribution of plausible poses~\cite{biggs2020multibodies, xu2020ghum,zanfir2020weakly}. These priors are usually trained using unpaired MoCap data~\cite{mahmood19amass}. Our work is fundamentally different from these methods in the sense that we are interested in learning a pose prior \emph{conditioned} on 2D image evidence rather than a generic prior on the 3D pose space.

\section{Method}
\begin{figure*}[!htb]
\vspace{-0.05in}
	\includegraphics[width=\textwidth,trim={0 14.5cm 15cm 0},clip]{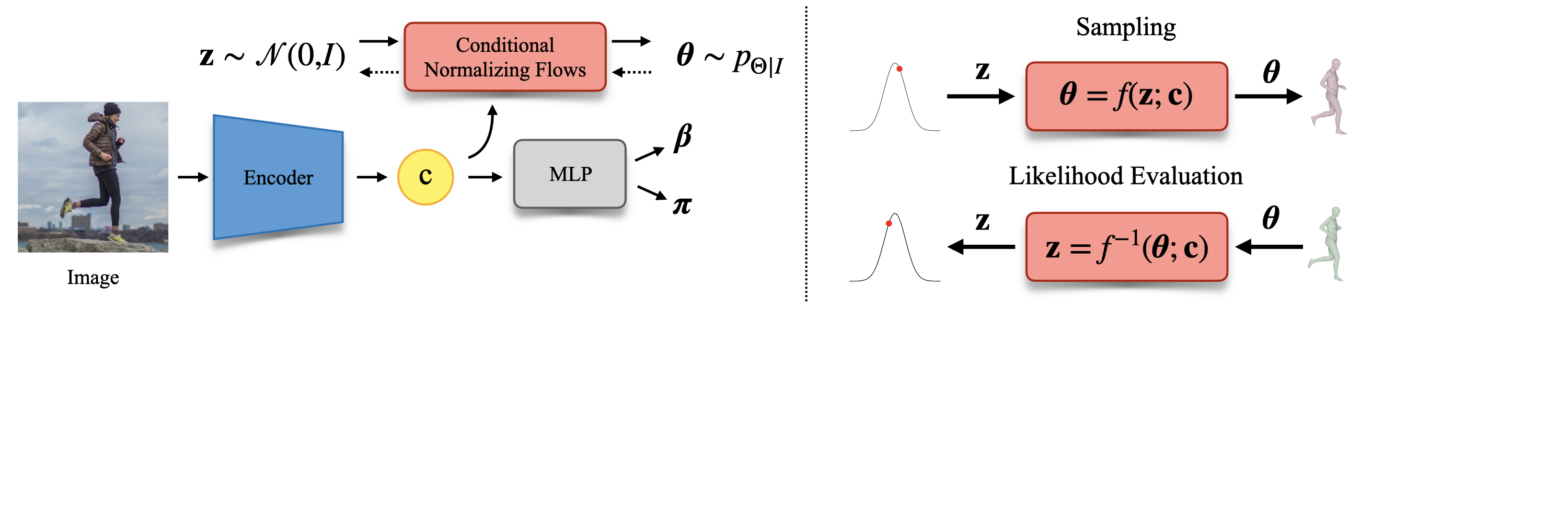} 
	\vspace{-0.3in}
	\caption{\textbf{Architecture of the proposed probabilistic model for human mesh recovery, ProHMR}. Left: Our image encoder regresses a hidden vector $\mathbf{c}$, which is used as the conditioning input to the flow model. In parallel, it is also decoded to shape parameters $\boldsymbol{\beta}$ and camera $\boldsymbol{\pi}$. Right: Our flow model learns an invertible mapping which allows for two processing directions; depending on the desired function, we can perform both sampling and fast likelihood computation.}
	\label{fig:method}
	\vspace{-0.2in}
\end{figure*}
In this Section, we present in detail our proposed approach. We start with an outline of Normalizing Flows~\cite{rezende15flows} and the SMPL body model~\cite{loper2015smpl}.  Then, we describe the model architecture and the training procedure. Finally, we show how our trained model can be used in downstream applications in a simple and straightforward manner.

\subsection{Normalizing Flows}
Let $Z \in \mathbb{R}^d$ be a random variable with distribution $p_Z(\mathbf{z})$  and $f : \mathbb{R}^d \to \mathbb{R}^d$ an invertible mapping. If we transform $Z$ with $f$, then the resulting random variable $X = f(Z)$ has probability density function:
\begin{equation}
p_X(\mathbf{x}) = p_Z(\mathbf{z})	\, \left\vert\operatorname{det}\frac{\partial f}{\partial \mathbf{z}}\right\vert^{-1}
\end{equation}

Normalizing Flow models are used to model arbitrarily complex distributions as a series of invertible transformations of a simple base distribution. Typically, the base distribution $p_Z(\mathbf{z})$ is chosen to be the standard multivariate Gaussian $\mathcal{N}(\mathbf{0}, I)$. If we write $f$ as a composition of invertible transformations $\{f_k\}_{k=1}^K$ with $Z_{0} = Z$, $Z_{i} = f_i(Z_{i-1})$ and $Z_K=X$, then the log-probability density of $X$ can be computed as:
\begin{equation}
\ln p_X(\mathbf{x}) = \ln p_Z(\mathbf{z})	-\sum_{k=1}^K \ln \left\vert\operatorname{det}\frac{\partial f_i}{\partial {\mathbf{z}}_{i-1}}\right\vert.
\end{equation}

Winkler~\etal~\cite{winkler2019learning} extended Normalizing Flow models to model conditional distributions $p_{X|Y}(\mathbf{x} \vert \mathbf{y})$ by using transformations $\mathbf{x}=f(\mathbf{z};\mathbf{y})$ that are bijective in $\mathbf{x}$ and $\mathbf{z}$.

\subsection{SMPL model}

SMPL~\cite{loper2015smpl} is a parametric human body model. It defines a mapping $\mathcal{M}(\boldsymbol{\theta}, \boldsymbol{\beta})$ that takes as input a set of pose parameters $\boldsymbol{\theta}$ and shape parameters $\boldsymbol{\beta}$ and outputs a body mesh $M \in \mathbb{R}^{N \times 3}$, where $N=6890$ is the number of mesh vertices. Additionally, given an output mesh, the body joints $J$ can be expressed as a linear combination of the mesh vertices, $J= WM$, where $W$ is a pretrained linear regressor.

\subsection{Model design}

Without loss of generality, we present our pipeline for the case where the input is an image of a person and the target output is the set of SMPL body model parameters. We call this model \textbf{ProHMR}, with the goal of \emph{Probabilistic Human Mesh Recovery}. At the end of this section we also show how the same method can be applied in alternative scenarios with different input and output representations.

In our setting, we are given an input image $I$ containing a person, and our goal is to learn a distribution of plausible poses for that person conditioned on $I$. Since we do not have access to accurate pairs of images-shape annotations, we choose to only model the uncertainty of the SMPL pose parameters $\boldsymbol{\theta}$. Our architecture follows closely the HMR paradigm~\cite{kanazawa2018end}. The output of our network is the conditional probability distribution $p_{\Theta \vert I}(\boldsymbol{\theta} | I)$ as well as point estimates for the shape and camera parameters $\boldsymbol{\beta}$ and $\boldsymbol{\pi}$ respectively.

The complete pipeline is depicted in Figure~\ref{fig:method}. 
Given an input image $I$, we encode it using a CNN $g$ and obtain a context vector $\mathbf{c} = g(I)$.  We model $p_{\Theta \vert I}(\boldsymbol{\theta} | \mathbf{c}=g(I))$ using Conditional Normalizing Flows. We learn a mapping $f: \mathbb{R}^d \times \mathbb{R}^c \to \mathbb{R}^d$ that is bijective in $\mathbf{z}$ and $\boldsymbol{\theta}$, \ie, $\boldsymbol{\theta} = f(\mathbf{z}; \mathbf{c})$ and $\mathbf{z} = f^{-1}(\boldsymbol{\theta}; \mathbf{c})$.

We employ Normalizing Flows instead of simpler alternatives such as Mixture Density Networks (MDN)~\cite{li2019cvpr} because of their expressiveness and ability to model more complex distributions, as we show later in the evaluation section. In our setting, Normalizing Flows have also clear advantages over VAEs, since VAEs do not offer an easy way to compute the likelihood of a given output sample, which is crucial when using our model in downstream tasks.

Our Normalizing Flow model is based on the Glow architecture~\cite{kingma2018glow}. Each building block $f_i$ is comprised of 3 basic transformations:
\begin{equation}
f_i = f_{coupl} \circ f_{lin} \circ f_{norm},
\end{equation}
where $f_{norm}(\mathbf{z}) = \mathbf{a} \odot \mathbf{z} + \mathbf{b}$ (Instance Normalization), $f_{lin}(\mathbf{z}) = W \mathbf{z} + \mathbf{b}$ (Linear transformation) and $f_{coupl} = [\mathbf{z}_{1:k}, \mathbf{z}_{k+1:d} + \mathbf{t}(\mathbf{z}_{1:d}, \mathbf{c})]$ (Additive coupling). To make the inversion and the Jacobian computation faster, in the linear transformation we parametrize the $LU$ decomposition of $W$. The final flow model is obtained by composing four of these building blocks.

The selected flow model allows us to perform both fast likelihood computation and fast sampling from the distribution. At the same time, a very important property is that the determinant of the Jacobian does not depend on $\mathbf{z}$, which in turn means that the mode of the output distribution is:
\begin{equation}
\boldsymbol{\theta}_I^\ast = \operatorname{argmax}_{\boldsymbol{\theta}} p_{\Theta \vert I}(\boldsymbol{\theta} \vert \mathbf{c}) = f(\mathbf{0}; \mathbf{c}).
\end{equation}
This result allows us to use our model as a \emph{predictive model} in a straightforward way; in the absence of any additional side-information, we make predictions using the mode of the output distribution.

To regress the camera and the SMPL shape parameters, we use a small MLP $h$ that takes as input the context vector~$\mathbf{c}$ and outputs a single point estimate, \ie, $[\boldsymbol{\beta}, \boldsymbol{\pi}] = h(\mathbf{c})$. We also experimented with having $\boldsymbol{\beta}$ and $\boldsymbol{\pi}$ depend on $\boldsymbol{\theta}$, but there was no observable improvement.

\subsection{Training objective}
Let us assume that we have a collection of images paired with SMPL pose annotations. Typically, Normalizing Flow models are trained to minimize the negative log-likelihood of the ground truth examples $\boldsymbol{\theta}_{gt}$, \ie the loss function is:
\begin{equation}
L_{nll} = - \ln p_{\Theta \vert I}(\boldsymbol{\theta}_{gt} \vert \mathbf{c}).
\end{equation}

However, for the task of 3D pose estimation, 3D annotations are generally not available except for a small number of indoor datasets captured in constrained studio environments~\cite{ionescu2014human3, mehta2017monocular} and methods trained on those datasets fail to generalize in challenging in-the-wild scenes. Consequently, previous methods like~\cite{kanazawa2018end} propose to use examples with only 2D keypoint annotations and minimize the keypoint reprojection loss jointly with an adversarial prior. To make such a mixed training possible within our framework, we propose to minimize the expectation of the above error with respect to the learned distribution, \ie,
\begin{equation}
L_{exp} = \mathbb{E}_{\boldsymbol{\theta} \sim p_{\Theta \vert I}}\lbrack  L_{2D}(\boldsymbol{\theta}, \boldsymbol{\beta}, \boldsymbol{\pi})  + L_{adv}(\boldsymbol{\theta}, \boldsymbol{\beta})\rbrack.
\end{equation}
To make this loss differentiable we use the \emph{Law of the Unconscious Statistician} and rewrite the expectation as:
\begin{equation}
L_{exp} = \mathbb{E}_{\mathbf{z} \sim p_{Z}}\lbrack L_{2D}(f(\mathbf{z}; \mathbf{c}), \boldsymbol{\beta}, \boldsymbol{\pi}) + L_{adv}(f(\mathbf{z}; c), \boldsymbol{\beta})\rbrack.
\end{equation}

Conceptually, even though we do not have ground truth annotations, to maximize the conditional probability of these examples we can still constrain the form of the output distribution by forcing the output samples to have low reprojection error on average and lie on the manifold of valid poses. As in the case of VAEs~\cite{kingma2014auto}, we approximate the expectation by drawing a single sample from the prior.

As mentioned previously, our goal is to use our model not only as a generative model but also as a predictive model. Thus, we propose to exploit the property that for each image $I$, the mode $\boldsymbol{\theta}_I^\ast$ of the output distribution corresponds to the transformation of $\mathbf{z} = \mathbf{0}$. We do this by explicitly supervising $\boldsymbol{\theta}_I^\ast$ with all the available annotations as in a standard regression framework and minimize:
\begin{equation}
L_{mode} = L_{3D}(\boldsymbol{\theta}_I^\ast, \boldsymbol{\beta})  + L_{2D}(\boldsymbol{\theta}_I^\ast, \boldsymbol{\beta}, \boldsymbol{\pi})  + L_{adv}(\boldsymbol{\theta}_I^\ast, \boldsymbol{\beta}),
\end{equation}
where $L_{3D}$ is the loss on the available 3D annotations (3D joints and/or SMPL parameters) whenever they are available. As we show in the experimental section, this explicit supervision of the mode of the output distribution helps boost the performance of our model in predictive tasks.

It is important to mention that $L_{exp}$ is not redundant in the presence of $L_{mode}$; the behavior of the mode is not indicative of the full distribution, whereas $L_{exp}$ encourages the distribution to have certain desirable properties.

Finally, for modeling rotations we use the 6D representation proposed in \cite{zhou2018continuity}. One issue with this particular representation is that it is not unique. For example, for any 3D vectors $x$ and $y$, $[x, y]$ and $[\alpha x, \beta x + \gamma y]$ are mapped to the same rotation matrix. Empirically we found that putting no constraints on the 6D representation results in large discrepancy between examples with full 3D SMPL parameter supervision and examples with only 2D keypoint annotations. Among other things, this caused mode collapse for the examples without 3D ground truth. Thus, we introduce another loss function $L_{orth}$ that forces the 6D representations of the samples drawn from the distribution to be close to the orthonormal 6D representation.

Eventually, the final training objective becomes:
\begin{equation}
L = \lambda_{nll} L_{nll} + \lambda_{exp} L_{exp} + \lambda_{mode} L_{mode} + \lambda_{orth} L_{orth}.
\end{equation}

\subsection{Downstream applications}
In this part we show how our learned conditional distribution can be used in a series of downstream applications. We highlight that all these applications refer to \emph{test-time} processing with the same trained model without any special per-task training. Examples of such tasks are shown in \figurename~\ref{fig:downstream}. These applications fall under the more general umbrella of Maximum a Posteriori estimation where we use all available evidence to make more informed predictions.

\paragraph{3D pose regression}
As already discussed in previous sections, we can use our model in conventional tasks such as 3D pose regression from a single image.
In the absence of additional evidence, the most appropriate choice for making predictions is to pick the mode $\boldsymbol{\theta}_I^\ast$ of the distribution.

\paragraph{Body model fitting}
SMPLify~\cite{bogo2016keep} is a popular method that fits the SMPL body model to a set of 2D keypoints using a traditional optimization approach. The objective is:
\begin{equation}
\lambda_J E_J + \lambda_\theta E_{\theta} + \lambda_\alpha E_\alpha + \lambda_\beta E_\beta,
\end{equation}
where $E_J$ penalizes the weighted 2D distance between the projected model joints and the detected joints, $E_\theta$ is a Mixture of Gaussian 3D pose prior, $E_\alpha$ is a pose prior penalizing unnatural rotations of elbows and knees and $E_\beta$ is a quadratic penalty on the shape coefficients.

Fitting a parametric body model to 2D image landmarks is a very challenging and inherently ambiguous problem. The data term $E_J$ is purely driven by the 2D keypoints and disregards rich information contained in the input image. SPIN~\cite{kolotouros2019spin} partially addresses this issue by using an image-based regression network that provides a good initialization for the optimization, helping the fitting to converge to a better minimum. However, the image information is only used in the initialization phase, as SMPLify does not incorporate explicit image-specific priors that prevent the pose to deviate arbitrarily far from the set of plausible poses for the given image. The drifting problem is also an important limitation of~\cite{joo2020eft}, forcing the approach to rely on good initialization and carefully chosen stopping criteria.

Motivated by these limitations, we propose to replace the weaker generic 3D priors $E_\theta$ and $E_\alpha$ with an explicit pose prior $E_{\theta \vert I } = -\ln p_{\Theta \vert I}(\boldsymbol{\theta} \vert \mathbf{c})$ that models the likelihood of a given pose conditioned on the image evidence. Thus, the final optimization objective becomes:
\begin{equation}
\lambda_J E_J -\ln p_{\Theta \vert I}(\boldsymbol{\theta} \vert \mathbf{c}) + \lambda_\beta E_\beta.
\end{equation}
As initialization for the fitting we use the mode $\boldsymbol{\theta}_I^\ast$ of the conditional distribution. In the experimental section we show that by using this learned image-based prior we are able to consistently improve the fitting results, both qualitatively and quantitatively, as reflected in the 3D metrics.

\paragraph{Multiple views fusion}
Although our model has been trained for single-image reconstruction, we can still use the learned conditional distribution to obtain refined pose estimates in the presence of multiple views of a person. Let us assume that we have a set $\{I_n\}_{1}^N$ of uncalibrated views of the same subject. We partition the pose vector of each frame as $\boldsymbol{\theta}_n = (\boldsymbol{\theta}_n^g, \boldsymbol{\theta}_n^b)$ where $\boldsymbol{\theta}_n^g$ corresponds to the global rotation of the model and $\boldsymbol{\theta}_n^b$ is the body pose. We propose to refine the pose by minimizing the following objective:
\begin{equation}
- \sum_{n=1}^{N} \ln p(\boldsymbol{\theta}_n | \mathbf{c}_n) + \lambda \sum_{n=1}^N \vert\vert \boldsymbol{\theta}_n^b - \bar{\boldsymbol{\theta}}^b\vert\vert_2^2,
\end{equation}
where $\bar{\boldsymbol{\theta}}^b = \frac{1}{N} \sum_{n=1}^N \boldsymbol{\theta}_n^b$. The second term of the objective is equivalent to minimizing the squared distance between all pairs of poses.

\subsection{Additional details}
\noindent
\textbf{ProHMR}.
Following previous works~\cite{kanazawa2018end, kolotouros2019spin} we use ResNet-50~\cite{he2016deep} as the encoder. For the Normalizing Flows we use 4 building blocks $f_i$. For more details about the architecture, datasets and the training hyperparameters we refer the reader to the supplementary material.

\noindent
\textbf{2D pose lifting}.
Complementary to ProHMR, we use our approach to lift 2D poses to 3D skeletons, as in Martinez~\etal~\cite{martinez2017simple}. We use the same Normalizing Flow architecture as in ProHMR. In this case the input is a set of 2D Hourglass detections~\cite{newell2016stacked} and the output is the 3D pose coordinates. For the encoder $g$, instead of a CNN, we use the backbone from~\cite{martinez2017simple}. Since all examples have full 3D supervision, our training objective consists only of $L_{nll}$ and $L_{mode}$.

\noindent
\textbf{Downstream tasks}.
For the fitting procedure employed in the downstream tasks, we found it beneficial to perform the optimization in the latent space instead of the pose space directly (similarly to SMPLify-X~\cite{pavlakos2019expressive}). Thus, we leave $\mathbf{z}$ as a free variable and decode it into the pose vector $\boldsymbol{\theta} = f(\mathbf{z}; \mathbf{c})$. Also, since for our Normalizing Flow model the determinant of the Jacobian does not depend on $\mathbf{z}$, the likelihood term becomes $\ln p(\boldsymbol{\theta} \vert \mathbf{c}) = - \vert\vert\mathbf{z}\vert\vert_2^2 + \text{const}$.

\section{Experimental evaluation}
In this Section we present the experimental evaluation of our approach.
First we provide an outline of the datasets used for training and evaluation
and then we will present detailed quantitative and qualitative evaluation results.

\subsection{Datasets}
We report results on Human3.6M~\cite{ionescu2014human3}, MPI-INF-3DHP~\cite{mehta2017monocular}, 3DPW~\cite{von2018recovering} and Mannequin Challenge~\cite{li2019mannequin}, where we use the annotations produced by Leroy~\etal~\cite{leroy2020smply}. For training, we use datasets with 3D ground truth (Human3.6M~\cite{ionescu2014human3} and MPI-INF-3DHP~\cite{mehta2017monocular}), as well as datasets with 2D keypoint annotations (COCO~\cite{lin2014microsoft} and MPII~\cite{andriluka20142d}) augmented with pseudo ground truth SMPL parameters from SPIN~\cite{kolotouros2019spin}, whenever they are available.

\subsection{Quantitative evaluation}
In this part we evaluate different aspects of our proposed approach.
We compare the predictive accuracy of our model with standard regression methods and show that it achieves comparable performance with the state of the art in human mesh recovery.
We also benchmark the generative capabilities of our method in multiple hypotheses scenarios, where we outperform previous approaches.
Finally, we demonstrate that our learned image-conditioned prior can boost the performance in downstream applications such as model fitting and multi-view refinement.

\begin{figure*}[!t]
	\centering
		\vspace{-5mm}
	\includegraphics[width=\textwidth,trim={0 9.8cm 0cm 0},clip]{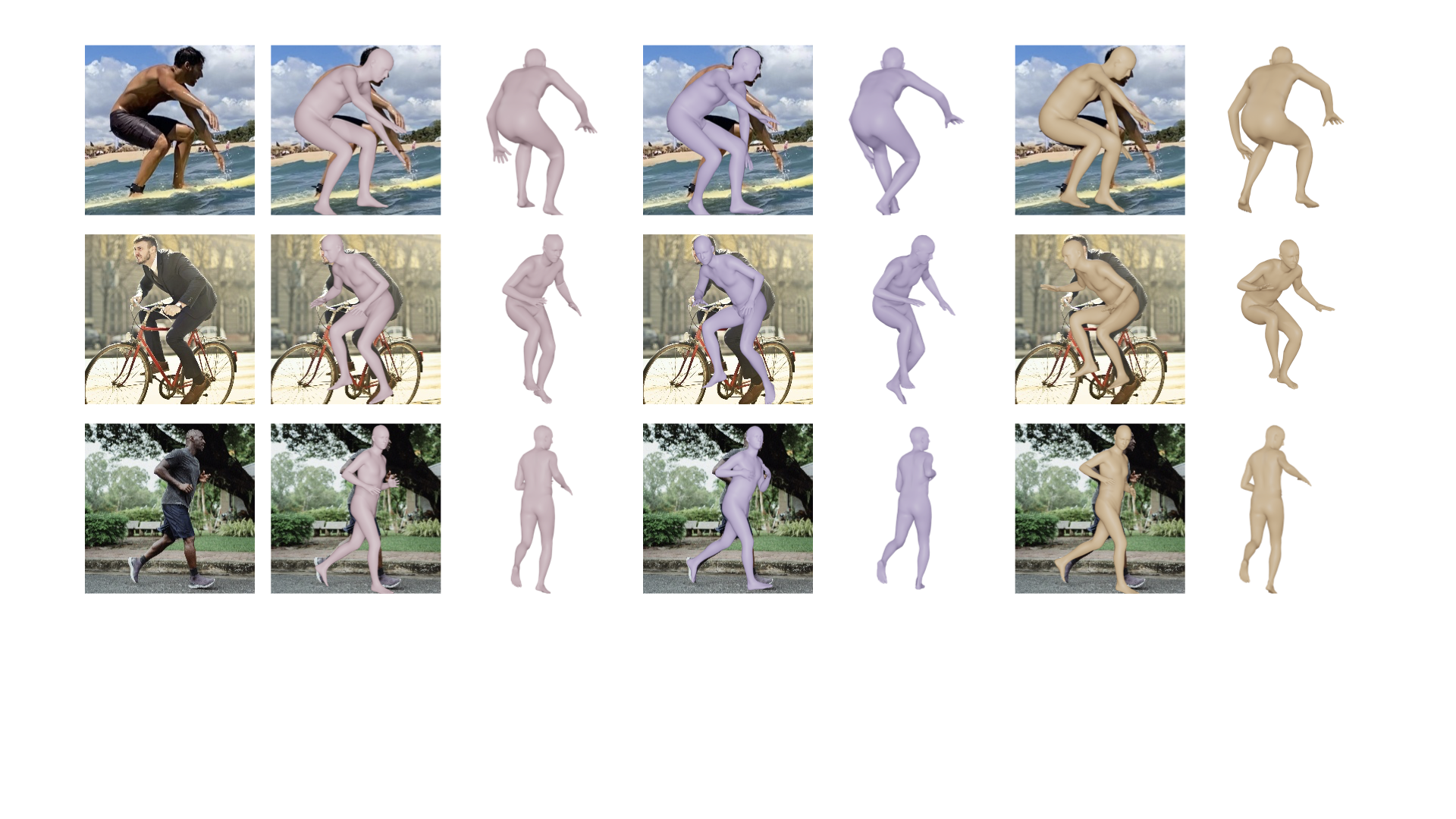} 	
		\vspace{-7mm}
	\caption{\textbf{Samples from the learned distribution}. Pink colored mesh corresponds to the mode.}
		\vspace{-4mm}
	\label{fig:qualitative_samples}
\end{figure*}

\noindent
\textbf{Human mesh recovery}. First, we focus on the predictive performance of our model, comparing it against other state-of-the-art methods that regress SMPL body model parameters. For the evaluation of ProHMR, we use the mode $\boldsymbol{\theta}_I^\ast$ of the learned distribution. For Biggs~\etal~\cite{biggs2020multibodies} we report the metrics after quantizing to $n=1$ sample. Based on the results of Table~\ref{tab:meshrecovery}, using ProHMR as a regressor, leads to performance comparable to the state of the art. This shows that we can indeed recast the problem from point to density estimation without any significant loss in performance.
\begin{table}
	\centering
	\small
	\hspace{-3mm}
	\begin{tabular}{ @{}lccc@{} }
		\toprule
		& 3DPW & H36M & MPI-INF-3DHP\\
		\midrule
		HMR~\cite{kanazawa2018end} & 81.3 & 56.8 & 89.8\\
		SPIN~\cite{kolotouros2019spin} & 59.1 & 41.1 & 67.5\\
		\midrule
		Biggs~\etal~\cite{biggs2020multibodies}  & 59.9 & 41.6 & N/A\\
		ProHMR & \textbf{59.8} & \textbf{41.2} & \textbf{65.0}\\
		\bottomrule
	\end{tabular}
	\vspace{-2mm}
	\caption{\textbf{Evaluation on human mesh recovery}. Our model achieves accuracy comparable with the state of the art. Numbers reported are PA-MPJPE in mm.}
	\label{tab:meshrecovery}
	\vspace{-4mm}
\end{table}

\noindent
\textbf{Multiple hypotheses}. Next, we compare the representational power of ProHMR with different multiple hypotheses baselines, including Biggs~\etal~\cite{biggs2020multibodies},  as well as the MDN and Conditional VAE variants explored in the same paper. Following \cite{biggs2020multibodies}, we report results for small sample sizes $n$. Since we are interested in measuring the representational power of the learned distribution,  we also compare the minimum 3D pose error of samples drawn from each distribution as proposed in~\cite{sharma19monocular}. We present the detailed results for Human3.6M and 3DPW in Table~\ref{tab:multihypotheses}. 

\begin{table}
	\centering
	\scriptsize
	\hspace{-3mm}
	\begin{tabular}{ @{}l|cc|cc|cc|c
	c@{} }
		& \multicolumn{2}{|c}{$n=5$} & \multicolumn{2}{|c}{$n=10$} & \multicolumn{2}{|c}{$n=25$} & \multicolumn{2}{|c}{$min$} \\
		\toprule
	    & \cite{von2018recovering} & \cite{ionescu2014human3} &  \cite{von2018recovering} & \cite{ionescu2014human3} &  \cite{von2018recovering} & \cite{ionescu2014human3} &  \cite{von2018recovering} & \cite{ionescu2014human3}\\
		\midrule
		\cite{biggs2020multibodies} (MDN) & 61.2 & 43.3 & 60.7 & 43.0 & 60.1 & 42.7  & 60.1 & 42.7\\
		\cite{biggs2020multibodies} (CVAE) & 60.7 & 46.4 & 60.5 & 46.3 & 60.3 & 46.2 & 60.3 & 46.2\\
		\cite{biggs2020multibodies} (NF) & 57.1 & 42.0 & 56.6 & 42.2 & 55.6 & 42.2 & 55.6 & 41.6\\
		ProHMR  & \textbf{56.5} & \textbf{39.4} & \textbf{54.6} & \textbf{38.3} & \textbf{52.4} & \textbf{36.8} & \textbf{40.8} & \textbf{29.9}\\
		\bottomrule
	\end{tabular}
	\vspace{-2mm}
	\caption{\textbf{Multiple hypotheses evalutation}. Numbers are PA-MPJPE in mm. We report errors for small $n$ and the \emph{minimum} error over samples drawn from the distribution.}
	\label{tab:multihypotheses}
\end{table}

\begin{figure*}[!t]
	\centering
		\vspace{-2mm}
	\includegraphics[width=\textwidth,trim={0 0cm 0cm 8.5cm},clip]{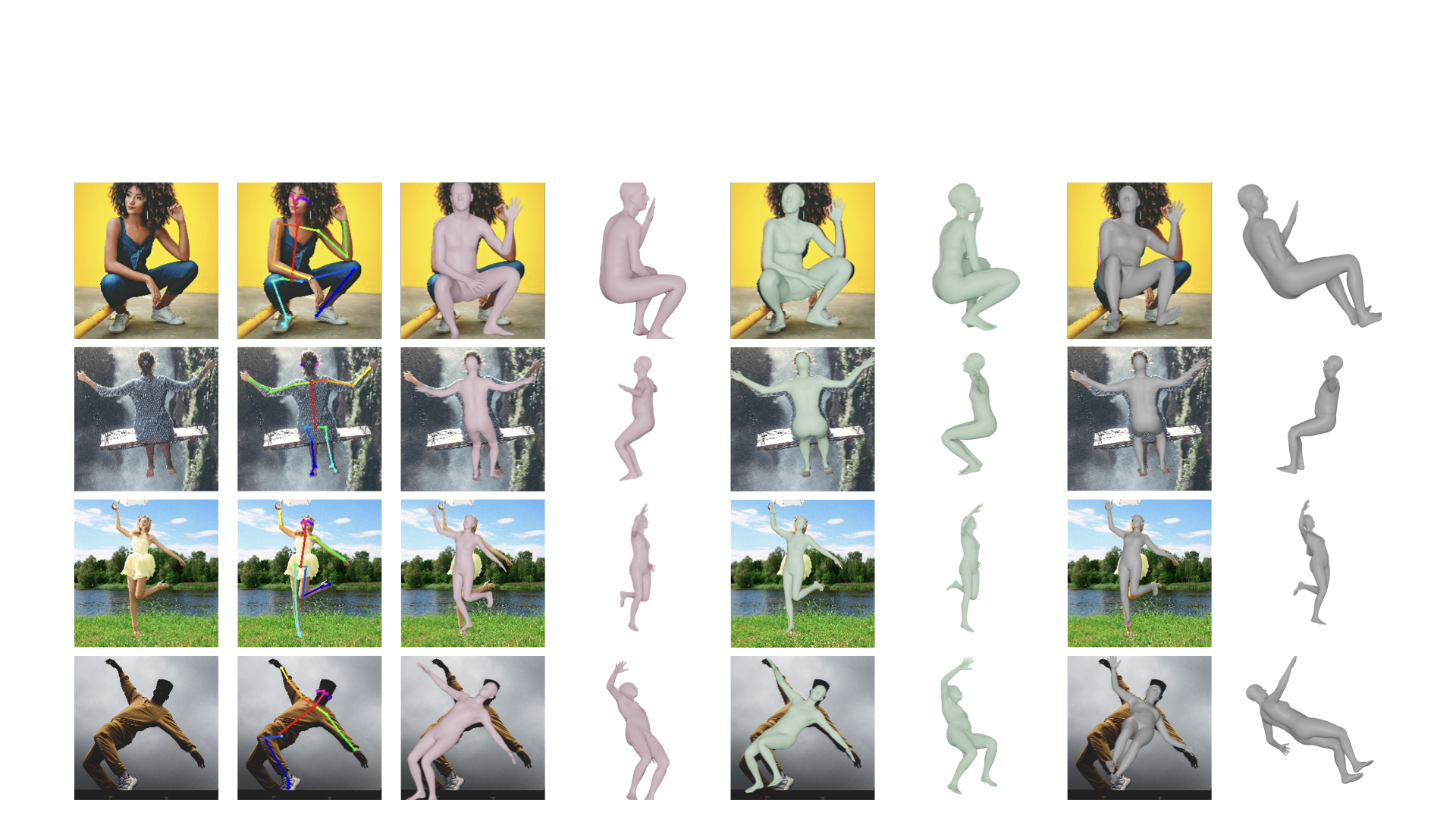} 	
		\vspace{-8mm}
	\caption{\textbf{Model fitting results}. Pink: Regression. Green: ProHMR + fitting. Grey: Regression + SMPLify}
		\vspace{-4mm}
	\label{fig:qualitative_smplify}
\end{figure*}

\noindent
\textbf{Model fitting}. In this part we evaluate the accuracy of different methods that fit the SMPL body model to a set of 2D keypoints. The body model fitting baselines we compare include the standard SMPLify~\cite{bogo2016keep,pavlakos2019expressive}, EFT~\cite{joo2020eft}, and our proposed fitting with the learned image-conditioned prior. For both SMPLify and EFT we use publicly available implementations and initialize the fitting process with SPIN, while for SMPLify we use two different versions for the pose prior, GMM~\cite{bogo2016keep} and VPoser~\cite{pavlakos2019expressive}. For a fair evaluation of the performance benefit, we compare methods that are trained on the same datasets and have similar regression performance. The results are presented in Table~\ref{tab:modelfitting}. While performing SMPLify on top of regression improves the model-image alignment, it increases the 3D pose errors, especially when using OpenPose detections~\cite{cao2018openpose}. We hypothesize that this happens because of the generic 3D pose prior terms of SMPLify. EFT on top of regression improves the 3D pose metrics, however our method manages to push the accuracy even further. In 3DPW our approach has a 4.7mm relative error improvement vs. 2.6mm for EFT, while if we use the ground truth 2D keypoints in Human3.6M we get a 6.3mm improvement vs 3.1mm for EFT.

\begin{table}
	\centering
	\small
	\hspace{-3mm}
	\begin{tabular}{ @{}lccc@{} }
		\toprule
		& 3DPW & H36M {\scriptsize (OP)} & H36M {\scriptsize (GT)} \\
		\midrule
		SPIN~\cite{kolotouros2019spin} & 59.2 & 41.8 & 41.8\\
		SPIN+SMPLify {\scriptsize (GMM)}~\cite{bogo2016keep} & 66.5 & 54.6 & 43.3\\
		SPIN+SMPLify {\scriptsize (VPoser)}~\cite{pavlakos2019expressive} & 70.9 & 53.5 & 39.9\\
		SPIN+EFT~\cite{joo2020eft} & 56.6  & 41.6 & 38.7\\
		\midrule
		ProHMR  & 59.8 & 41.2 & 41.2	\\
		ProHMR + fitting & \textbf{55.1}  & \textbf{39.3} &  \textbf{34.8} \\
		\bottomrule
	\end{tabular}
	\vspace{-2mm}
	\caption{\textbf{Evaluation of different model fitting methods.} The fitting algorithms are initialized by the corresponding regression results. All numbers are PA-MPJPE in mm. }
	\label{tab:modelfitting}
	\vspace{-4mm}
\end{table}

\noindent
\textbf{Multi-view refinement}. We evaluate the effect of our learned image-conditioned prior at refining the pose predictions in uncalibrated multi-view scenarios. For benchmarking, we use Human3.6M and the more challenging Mannequin Challenge dataset. We compare our fitting-based method against the individual per-view predictions and a baseline that performs rotation averaging in Table~\ref{tab:multiview}. For the rotation averaging we first average the per-view rotation matrices and then project them back to $SO(3)$ using SVD.

\begin{table}
	\centering
	\footnotesize
	\hspace{-3mm}
	\begin{tabular}{ @{}lcccc@{} }
		\toprule
		& \multicolumn{2}{c}{H36M}  & \multicolumn{2}{c}{Mannequin}\\
		\cmidrule{2-5}
		& MPJPE & PA-MPJPE & MPJPE & PA-MPJPE\\
		\midrule
		ProHMR & 65.1 & 43.7 & 176.0 & 91.9\\
		ProHMR + rot avg & 64.8 & 35.2 & 174.4 & 85.1\\
		ProHMR + fitting& \textbf{62.2} & \textbf{34.5} & \textbf{171.3} & \textbf{83.9}\\
		\bottomrule
	\end{tabular}
	\vspace{-2mm}
	\caption{\textbf{Evaluation of multi-view refinement.} We compare single-image 3D reconstruction with a baseline refinement using rotation averaging and the proposed optimization-based refinement scheme.}
	\label{tab:multiview}
	\vspace{-4mm}
\end{table}

\noindent
\textbf{Ablation study}.
We also assess the significance of the term $L_{mode}$ that we use to explicitly supervise the mode of the learned distribution. We report results for training ProHMR with and without this loss in Table~\ref{tab:ablation}. 
We can see that including $L_{mode}$ is crucial to achieve competitive performance in conventional regression tasks.
\begin{table}
	\centering
	\small
	\hspace{-3mm}
	\begin{tabular}{ @{}lccc@{} }
		\toprule
		& 3DPW & H36M & MPI-INF-3DHP\\
		\midrule
		ProHMR (w/o $L_{mode}$) & 67.4 & 54.8 & 76.5\\
		ProHMR & \textbf{59.8} & \textbf{41.2} & \textbf{65.0}\\
		\bottomrule
	\end{tabular}
	\vspace{-2mm}
	\caption{\textbf{Ablation for $L_{mode}$}. Numbers are PA-MPJPE.}
	\label{tab:ablation}
	\vspace{-4mm}
\end{table}

\noindent
\textbf{Additional evaluations}.
Finally, we show that the proposed modeling is general enough to handle different input and output representations. Here, we consider the setting of lifitng a 2D pose input to a 3D skeleton output~\cite{martinez2017simple} and present results in Table~\ref{tab:skeleton}. Our model performs on par with an equivalent regression approach~\cite{martinez2017simple}, while it outperforms the MDN method of Li and Lee~\cite{li2019cvpr}.

\subsection{Qualitative results}
In \figurename~\ref{fig:qualitative_samples} we show sample reconstructions of our method. Additionally, in \figurename~\ref{fig:qualitative_smplify} we show comparisons of our model fitting approach with SMPLify. Our method produces more realistic reconstructions overall, particularly in cases where there are missing or very low confidence keypoint detections. In cases like that (\eg, example of last row), our image-based prior, unlike SMPLify, does not let the pose deviate far from the image evidence.

\begin{table}
	\centering
	\small
	\hspace{-3mm}
	\begin{tabular}{ @{}lcc@{} }
		\toprule
		&MPJPE & PA-MPJPE\\
		\midrule
		Martinez~\etal~\cite{martinez2017simple}  & 62.9 & 47.7\\
		Li and Lee~\cite{li2019cvpr} (mode) & 64.5 & 47.8\\
		Ours & 62.9 & 47.6\\
		\midrule
		Li and Lee~\cite{li2019cvpr} (min) & 42.6 & 34.4\\
		Ours (min) & \textbf{42.4} & \textbf{32.9}\\
		\bottomrule
	\end{tabular}
	\vspace{-2mm}
	\caption{\textbf{Evaluation of 3D pose accuracy for skeleton-based 2D pose lifting} on Human3.6M. Top: Regression accuracy. Bottom: Minimum error of the distributions.}
	\label{tab:skeleton}
	\vspace{-4mm}
\end{table}

\section{Summary}
This work presents a probabilistic model for 3D human mesh recovery from 2D evidence. Unlike most approaches that output a single point estimate for the 3D pose, we propose to learn a mapping from the input to a distribution of plausible poses. We model this distribution using Conditional Normalizing Flows. Our probabilistic model allows for sampling of diverse outputs,  efficient computation of the likelihood of each sample,  and a fast and closed-form solution for the mode. We demonstrate the effectiveness of our method with empirical results in several benchmarks. Future work could consider extending our approach to other classes of articulated or non-articulated objects and potentially model other ambiguities like the depth-size trade-off.

\footnotesize
\noindent
{\bf Acknowledgements:} 
Research was sponsored by the following grants: ARO W911NF-20-1-0080, NSF IIS 1703319, NSF TRIPODS 1934960, NSF CPS 2038873,  ONR N00014-17-1-2093, the DARPA-SRC C-BRIC, and by Honda Research Institute. GP is supported by BAIR sponsors.

{\small
\balance
\bibliographystyle{ieee_fullname}
\bibliography{egbib}
}

\end{document}